\documentclass[letterpaper, 10 pt, conference]{ieeeconf}  

\IEEEoverridecommandlockouts                              

\usepackage{graphicx}

\usepackage{algorithm,algpseudocode}
\algdef{SE}[DOWHILE]{Do}{doWhile}{\algorithmicdo}[1]{\algorithmicwhile\ #1}%
\usepackage{xcolor}
\usepackage{dblfloatfix}
\usepackage{cite}
\usepackage{flushend}

\usepackage{subfig}
\usepackage{amsfonts}
\usepackage{gensymb}
\usepackage{amsmath,amssymb,amsfonts}
\usepackage{url}
\usepackage{etoolbox}
\usepackage{graphicx}
\usepackage{hyperref}

\usepackage{fancyhdr}
\usepackage{tikz}

\fancypagestyle{mystyle}{%
    \fancyhead[CO, CE]{
This paper has been accepted for publication at IEEE International Conference on Robotics and Automation (ICRA) 2024.
    }
}%

\title{\LARGE \bf
On Experimental Emulation of Printability and Fleet Aware Generic Mesh Decomposition for Enabling Aerial 3D Printing}

\author{Marios-Nektarios Stamatopoulos, Avijit Banerjee, and George Nikolakopoulos
\thanks{The Authors are with the Robotics and Artificial Intelligence Group, Department of Computer, Electrical and Space Engineering, Lule\r{a} University of Technology, 971 87 Lule\r{a}, Sweden}%
\thanks{Cooresrponing author's e-mail: \tt\small{marsta@ltu.se}}
}

\begin{document}

\maketitle
\begin{tikzpicture}[overlay, remember picture]
    \node[anchor=north,
          xshift=0.0cm,
          yshift=-0.15cm, align=center]
         at (current page.north)
         {\fontsize{8}{10}\selectfont This paper has been accepted for publication at IEEE International Conference on Robotics and Automation (ICRA) 2024.
};
\end{tikzpicture}

\thispagestyle{empty}
\pagestyle{empty}

\begin{abstract}
This article introduces an experimental emulation of a novel chunk-based flexible multi-DoF aerial 3D printing framework. The experimental demonstration of the overall autonomy focuses on precise motion planning and task allocation for a UAV, traversing through a series of planned space-filling paths involved in the aerial 3D printing process without physically depositing the overlaying material. The flexible multi-DoF aerial $3$D printing is a newly developed framework and has the potential to strategically distribute the envisioned 3D model to be printed into small, manageable chunks suitable for distributed 3D printing. Moreover, by harnessing the dexterous flexibility due to the $6$ DoF motion of UAV, the framework enables the provision of integrating the overall autonomy stack, potentially opening up an entirely new frontier in additive manufacturing. However, it's essential to note that the feasibility of this pioneering concept is still in its very early stage of development, which yet needs to be experimentally verified. Towards this direction, experimental emulation serves as the crucial stepping stone, providing a pseudo mockup scenario by virtual material deposition, helping to identify technological gaps from simulation to reality. Experimental emulation results, supported by critical analysis and discussion, lay the foundation for addressing the technological and research challenges to significantly push the boundaries of the state-of-the-art 3D printing mechanism.
\end{abstract}
\section{Introduction}
In recent years, the field of $3$D printing has witnessed significant advancements at an unprecedented pace, transforming it into a powerful tool for rapid prototyping. The application of the $3$D printing technology has transformed what was once relegated to the realm of science fiction into a tangible reality~\cite{karasik2019object,balletti20173d}. Its current impact spans across diverse industries, including construction, agriculture, healthcare, automotive, and aerospace, unveiling exciting prospects for the future. Notably, it holds potential for constructing full-scale infrastructures, such as emergency shelters and post-disaster relief accommodations in remote, harsh, and hard-to-reach environments exposed to extreme climates, unstructured terrain, and distant military locations \cite{xu2022robotics}. Conventional manufacturing and 3D printing demand substantial machinery and infrastructure, limiting their utility in far-flung and inaccessible regions\cite{al2018large,bazli20233d}, while also constraining the size of built objects based on machinery dimensions, which hinders scalable construction, particularly in areas lacking the necessary infrastructure \cite{zhao2022general}.

\begin{figure}[t]
    \centering
    \includegraphics[width=0.9\columnwidth]{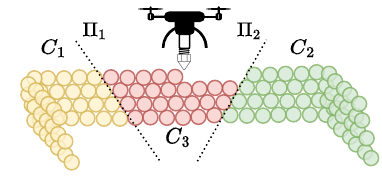}
    \setlength{\belowcaptionskip}{-18pt}
    \caption{Conceptual representation of aerial 3D printing UAV during the manufacturing process of chunk $C_3$. The whole construction is decomposed through planar cuts $\Pi_1$ and $\Pi_2$.}
    \label{fig:concept}
\end{figure}

By leveraging significant advancements in autonomous Unmanned Aerial Vehicle (UAV) technology, this visionary framework is emerging as a groundbreaking concept offering unparalleled flexibility, accessibility, and efficiency in autonomous and scalable manufacturing constructions. Nevertheless, aerial 3D printing is a pioneering concept still at its very early stage of development, with limited existing research, mostly in its conceptual phase~\cite{feasibility,kovac}. This article introduces an experimental emulation of a versatile and adaptable generic aerial 3D printing framework designed to decompose construction into individual manufacturing sub-tasks (as shown in Fig. \ref{fig:concept}), laying the foundation for more efficient and scalable construction operations, while it should be highlighted that both aspects in the field of material science and extruder technology are left out of the scope of this article as sequential future works.

\section{Related Work}
The context of multi-robot automated construction has been investigated mostly through ground robots equipped with manipulators~\cite{wenchao_zhou_ground_robots,Zhang2018,kanoulas}, but are limited to movements in a 2D plane and the reachability of the arm.
%
Aerial additive manufacturing is posed as an alternative to it, although it is still in an early stage.
Specifically, the approach  in~\cite{kovac} examined the feasibility and proof of concept of the hybrid of aerial robotics and additive layer manufacturing which primarily involves a single UAV in operation. In the context of constructing full large-scale constructions in hard-to-access areas, in~\cite{zhang2022aerial} the use of aerial platforms is examined where both scenarios with single and multiple UAVs are evaluated. The process of multi-UAV construction is executed through a conventional $2.5$D tactic, which entails the segmentation of the given mesh into discrete layers, subsequently dividing each layer into an equitably distributed set of curves, referred to as print jobs that are assigned to each UAV. 
For the multi-UAV scenario, a virtual printing environment was employed. Luminescent light markers were mounted on the UAVs, wherein alternations in their color would indicate the extrusion of the material. 
Towards the deployment of multiple UAVs for the realization of any given large-scale construction, an early-stage conceptual framework is presented in~\cite{stamatopoulos2023flexible} aiming to decompose any given mesh that is intended to serve as input for the subsequent manufacturing process. The decomposition is carried out in the form of an iterative search, while upon the generation of the sub-parts, they are sequentially assigned to each UAV. The framework is evaluated in Gazebo software by tracking the extruder tip and visualizing the material extrusion whenever the printing action is commanded from the UAV. 
\section{Contributions}
In this article, a novel, generic multiaxis chunk-based aerial 3D printing framework supported by experimental emulation is presented. In this context, a recently introduced chunk-based methodology~\cite{stamatopoulos2023flexible} is extended to incorporate an additional heuristic function in the iterative beam search process. This extension is strategically designed to increase the number of "seed" chunks, which hold significant importance due to their direct influence on the parallelizability of the manufacturing process. These seed chunks exhibit minimal manufacturing interdependencies, thereby facilitating the parallelization of printing tasks and bringing the framework closer to achieving the milestone of collaborative printing. In contrast with the previous state-of-the-art, the proposed framework enables seamless integration of distributed aerial 3D printing with multiple UAVs in a multiaxis layered manufacturing fashion. Moreover, the experimental emulation essentially considers the overall autonomy, incorporating precise coordinated motion planning of a UAV, while traversing a series of 
planned space-filling paths. It effectively utilizes a Model Predictive Control (MPC) strategy, with careful consideration of the unique attributes of the UAV model, ensuring necessary movements for successful execution. 
Experimental emulation results, focusing on the framework evaluation, motion planning excluding the material science and extruder technology are critically analyzed and discussed. Thus, paving the way and providing a foundational basis as a bridge between sim to real, identifying the technological challenges necessary to realize this innovative approach.
%

%
\section{Methodology} \label{Sec:Problem formulation}
\begin{figure*}[h]
    \begin{center}
        \includegraphics[width=\textwidth]{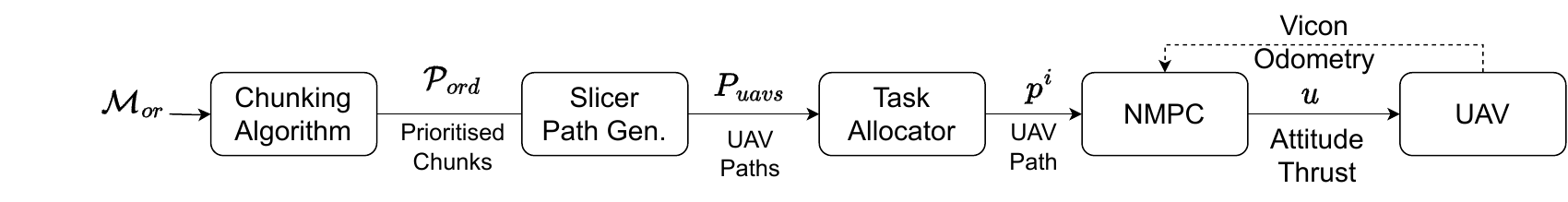}
        \setlength{\abovecaptionskip}{-10pt}
        \setlength{\belowcaptionskip}{-22pt}
        \caption{ {Framework Block Diagram} }
        \label{fig:UAVblock_diagram}
    \end{center}
\end{figure*}

%
The initial step in realizing aerial 3D printing involves an innovative optimization-based mesh decomposition approach. This process dissects the user-provided 
mesh into smaller segments called "chunks" by strategically selecting cutting planes. This selection employs a search-
based methodology \cite{stamatopoulos2023flexible}, wherein potential cutting planes are sampled and assessed. The most favorable combinations of planes 
advance to subsequent iterations, incorporating additional cuts.
To facilitate evaluation, a heuristic function is introduced maximizing volume dispersion among the chunks and enhancing 
the parallelability of chunk printing tasks. A Binary Space Partitioning Tree (BSP)~\cite{BSPTree} is established to 
track the planar cuts and resulting chunk structure. Post-search phase, the BSP tree acts as a task priority sequencer due to its inherent relationships within its structure, which dictate dependencies between the chunks. 
Subsequently, each chunk is allocated to a UAV in a responsive manner. Every chunk undergoes 
slicing through a commercially available slicer and is transformed into a trajectory, which the assigned UAV 
executes. An overview of this approach is depicted in Fig. \ref{fig:UAVblock_diagram}.

\section{Chunks Generation} \label{Sec:Chunk Generation}
%
The chunk generation process is carried out by intersecting the mesh with a set of planes, fragmenting it into smaller 
printable sub-parts. The cutting planes are denoted by $\Pi^j_i$, defined with the surface-normal vector 
$\Vec{n_i}\in R^3$, perpendicular to the plane, originated at $\Vec{p_j}\in R^3$ and laying on respective cutting 
surfaces. 
A BSP Tree $\mathcal{T}$ is utilized to keep track of the cuts and the resulting chunks. The original mesh is located at the root node, all the generated chunks $C_i$ are located at the leaves of $\mathcal{T}$, while all the cutting planes are in the rest of the nodes. When a new planar cut $\Pi_i$ is executed, a recursive process identifies all affected chunks $C_j$ and are extended into child nodes. Specifically, intermediate children nodes are labeled as $C^{-}{j,i}$ and $C^{+}{j,i}$, representing the outcome of the cut between $\Pi_i$ and $C_j$. Among these, $C^{+}{j,i}$ resides in the half-space aligned with the normal vector $\Vec{n_i}$ and is referred to as a positive chunk. Conversely, $C^{-}{j,i}$ is termed a negative chunk. This article follows the convention of positioning the negative chunk, $C^{-}_{j,i}$, as the left child of the parent node.

Towards finding the optimal combination of 
planar cuts $\Pi^j_i$  that will
result into generated chunks with specified characteristics to facilitate aerial collaborative printing, an iterative beam-search approach~\cite{mit_chopper} is employed.
%
%
%
%

Initially, normal vectors are uniformly sampled within a spherical space subject to constraints in the polar angle $\phi_{s} \in [-\phi_{s}^{max}, \phi_{s}^{max}]$, while $r=1$, and $\theta_{s} \in [0,2\pi)$.
A family of planes is calculated for each sampled normal vector along its axis with a predefined distance offset $\delta \in \mathbb{R}$ between them.
Each planar cut $Pi^j_i$ of the aforementioned sampling is applied to the corresponding tree $\mathcal{T}_i$ that requires expansion. Every newly generated tree $\mathcal{T}_{i,j}$, resulting from this extension process, is collected and in the sequel evaluated through a heuristic function. The top $W_{inner}$ ones are selected to be added to the set containing all the new trees generated in the current expansion iteration. After the evaluation of the new cuts is executed for all $\mathcal{T}_i$ trees, the top $W_{outter}$ ones are selected to proceed to the next iteration.


\subsection{Heuristics}
Each BSP tree containing both the optimal cuts and the chunks, needs to be evaluated and be given a cost value in order to compare it among the rest and keep the best ones for the next round of expansions. For this purpose, a heuristic function is assigned incorporating two different sub-heuristics. Uniform volume distribution of the generated chunks along with increase in the manufacturing parallelability of the structure are targeted.


\subsubsection{Volume Dispersion}
Uniformity in volume yields an evenly distributed arrangement of the constituent parts, with their geometric centers adequately spaced apart. This uniformity is translated into reduced waiting times between printing operations, as each constituent element's printing duration closely aligns with that of its counterparts.
Thus, the term volume dispersion $c_v ={\sigma} / {\mu} $ is used as heuristic, where $\mu=\sum_{i=1}^{N} V_i/{N}$, $\sigma = {\sum_{i=1}^{N} (V_i - \mu)^2}/{N}$ and $V_i$ being the volume of the $i$-th chunk.

%

\subsubsection{Seed Chunks}

In accordance with the notation established in~\cite{chunking_notation}, a "seed chunk" is precisely delineated as a chunk for which all resultant faces from the planar cuts exhibit positivity.
The maximization of the presence of such chunk types within the final hierarchical structure is of great importance, 
given that their sole prerequisite is to be fabricated either at ground level or atop a flat chunk. 
In this configuration, all of their faces act as supports for subsequent chunks within the sequence. This particular attribute enhances the potential for parallel execution of printing tasks, as it eliminates any additional inter-dependencies.
To determine the seed status of a given chunk, denoted as 
$C_k$, a set $\mathbf{C}=\{\Pi_i , \: 
i=1,2,\dots, M\}$ is assembled, where each $\Pi_i$ is co-planar with it. 
Via set $\mathbf{C}$ it is verified
whether the chunk aligns positively or negatively with the corresponding plane $\Pi_i$. 
This determination is achieved by the projection of an arbitrary 
vertex  $\Vec{v^k_j} \in \mathbb{R}^3$, from the $C_k$ mesh, onto the normal vector $\Vec{n_i}$ of $\Pi_i$. It is calculated according to $proj_{\Vec{n_i}}(\Vec{v^k_j}) = (\Vec{v^k_j} - \Vec{p_i}) \cdot \Vec{n_i}$
where, $\Vec{p_i}$ represents the origin of plane $\Pi_i$.
The final classification is determined by the value $s_k \in \mathbb{R}$, defined as follows:

\begin{equation}
    s_k = 
    \begin{cases}
        1 \quad &\text{if } proj_{\Vec{n_i}}(\Vec{v^k_j}) < 0, \:  \forall \: \Pi_i \in \mathbf{C} \\
        0 \quad &\text{otherwise} \\
        \end{cases}
\end{equation}

Hence, a chunk $C_k$ qualifies as a seed whenever $s_k = 1$. This attribute is utilized by including the term $s_k$ in the heuristic function. Furthermore, an extension is introduced where the count of positive faces of a chunk is rewarded, provided that it is classified as a seed. A positive face of a seed chunk $C_k$ is defined as a face that shares a co-planar alignment with a plane $\Pi_i$ and is represented by the term $f^{k+}_{i,j}$ being equal to $1$, having $proj_{\Vec{n_i}}(\Vec{v^k_j}) < 0$ as a positivity condition.
%
%
Consequently, the term ${p}^k_{faces} = \sum_{i=1}^{M} f^{k+}_{i,j} \: s_k $ denoting their number, is introduced as a reward in the heuristic. Both of the previously mentioned heuristics are integrated into the unified heuristic function:
\begin{equation}
    g(C_k) = G_{part} \: s_k + G_{faces} \: {p}^k_{faces} , C_k \in \mathcal{T} 
\end{equation}
where $G_{part}, G_{faces} \in \mathbb{R}$, gains adjusted to align with the goals of the particular application.

\subsection{Combined Heuristic}
Fusing the aforementioned heuristics in a single one, the total heuristic function $h(\mathcal{T})$, for the evaluation of the tree $\mathcal{T}$, is given by:
\begin{equation}\label{huristic_eq}
    h(\mathcal{T}) = G_{disp} c_v - \sum_{k=1}^{M} g(C_k)
\end{equation}
where $M$ represents the number of chunks within the specific BSP tree denoted as $\mathcal{T}$. The summation of functions $g(C_k)$ for each chunk is negated to stress the necessity of rewarding these values rather than imposing penalties.

\subsection{Printability Constraints}

During the phase of normal vector sampling, the polar angle $\phi$ of the sampling surface $\mathbf{S}$, and consequently, the inclination of each planar cut, is constrained by considerations related to both chunk inter-connectivity and potential extruder collisions~\cite{stamatopoulos2023flexible}. Specifically, an upper angular limit $\phi^{conn}_{max}$ is imposed to ensure the creation of adequate printable overlap regions between adjacent surfaces. This overlap is critical for providing the requisite support and eventual adhesion of the printed parts. Additionally, the extruder's geometric characteristics, approximated as a bounding rectangle, are considered to prevent any collisions with previously printed components. This consideration results in the determination of a maximum allowable cutting slope, denoted as $\phi^{extr}_{max} \leq \tan^{-1}\left(\frac{h}{l}\right) $. Finally, a
unified upper bound, expressed as $\phi_{max} = max(\phi^{conn}_{max}, \phi^{extr}_{max}) $ is used.

\subsection{Tree Terminal Condition}
The initial UAVs' setup is considered known before the chunking phase. Specifically, the volume of the carried extrusion material is given as input to the algorithm. Both UAVs' carrying material $d^n_{m}$ and chunks volumes $c_k$ are sorted in descending order and placed in the sets $\mathbf{D_{mat}} = [d^1_{m}, d^2_{m}, \dots, d^n_{m}]$ and $\mathbf{C}=[c_1,c_2, \dots,c_k]$ for the UAVs and chunks correspondingly. 
%
%
A tree is considered to be terminated when all the chunks generated at its leaf nodes can be printed by at least one of the available UAVs. Hence, the algorithm stops searching for further cuts to extend it when a non-empty set exists and is defined as:
\begin{eqnarray} \label{eq:term_condition}
    S = \{p_l= (d_j,c_i) :  d_j>c_i\} \neq \O, \forall l=[1,\dots,k]
\end{eqnarray}
The sets $C, D$ and $S$ are updated recursively after the construction of each chunk.

\section{UAV EXECUTION} \label{Sec:UAV Execution}
Once a chunk is designated for printing, a prescribed sequence of actions must be executed to initiate the printing process. 
Each chunk mesh is given to the slicer that is configured with precise parameters including UAV and extruder dimensions and material deposition variables, adjusted to suit the project's specific requirements. It is transformed into the adequate G-code commands for effectively manufacturing the chunk. A block diagram depicting the sequential workflow associated with this process is illustrated in Fig. \ref{fig:UAVblock_diagram}.

\subsection{G-Code to UAV Path}
A post-processing step is necessary to eliminate extraneous data from the G-code and to convert the 
path into a simplified format compatible with the UAV. All G-code segments are transformed into 
equivalent waypoints that the UAV must traverse, resulting in a path $P^i_{ex}$ for the $i$-th chunk. This path
corresponds to the extruder's path, not the UAV's. To bridge this gap, a transformation converting the extruder tip's frame to the UAV's body frame is applied. To facilitate this 
transformation, an extruder of a length $l_{ex}$ is employed for the printing operation, with the extruder tip vertically oriented toward the ground. Finally, the path $P^i_{uav}$ referring to the $i$-th chunk is obtained.

\subsection{Task Allocation}
The BSP tree organizes planar cuts during the search phase, with chunks allocated to its leaves for orderly mesh creation. Constraining the plane's slope with $\phi_{max}$ ensures a positive vertical component (along $z$ axis) in normal vectors $\Vec{n_i}$. This results in a dependency between the two resulting chunks after a cut: the positive chunk $C^{+}{j,i}$ lies above the plane $\Pi_i$, while the negative chunk $C^{-}{j,i}$ below.
Strategically selecting the preceding printed chunks as supports for the following ones, the left child nodes and their subordinates in the BSP tree $\mathcal{T}$ take precedence over the right ones. To eliminate dependency on plane selection order, planar cuts in the BSP tree $\mathcal{T}$ are reordered based on origin point $\Vec{p}$ $z$-coordinates, from low to high. After executing the cuts with the aforementioned order, the same priority rule is applied and the priority order set $\mathcal{P}_{ord}$ is generated by traversing the tree in-order. $\mathcal{P}{ord}$ comprises the requisite sequence in which the individual chunks must be printed to facilitate the construction in a feasible manner.

Throughout the execution of the printing process, the allocation of each chunk to the UAVs is conducted in a reactive fashion. The chunks are printed sequentially following the rule that each chunk must be printed by only one UAV. Whenever a UAV is in an idle state, no other UAV is printing at the time and there are still chunks pending to be printed, it is assigned the next chunk in order and the corresponding chunk is excluded from the $\mathcal{P}_{ord}$ set becoming no longer available. In the case that a UAV is running low on battery, it sets itself to an inactive state and is no longer able to get assigned chunks to print. The aforementioned process is executed iteratively until there are no more chunks to be manufactured.

\subsection{UAV Motion Planning Framework}
For precise execution of aerial 3D printing, each UAV must follow the designated path 
$P_{UAV}$, while ensuring uniform material deposition. To achieve this, a Nonlinear Model 
Predictive Control (NMPC)-based optimal path following controller~\cite{bjorn_mpc} is employed 
and designed to accurately track the path generated by the G-code. The UAV's dynamics are 
described using two coordinate frames: 
the world frame $W\in \mathbb{R}^3$ 
and the body-fixed frame $B\in \mathbb{R}^3$. 
The body-fixed frame is 
centered on the UAV's mass center, while the inertial frame is assumed to be fixed in the origin. The UAV's motion is governed by a set of non-linear dynamic equations presented as:
\begin{eqnarray}
\dot{p}(t) &=& v(t) \nonumber\\
\dot{v}(t) &=& R_{x,y}(\theta,\phi) \mathbf{T} + \mathbf{G} - \mathbf{A} v(t) \label{eq:quadcopter_dynamics}\\
\dot{\phi}(t) &=& \frac{1}{\tau_{\phi}} (K_{\phi} \phi_{ref}(t) - \phi(t)) \nonumber\\
\dot{\theta}(t) &=& \frac{1}{\tau_{\theta}} (K_{\theta} \theta_{ref}(t) - \theta(t)) \nonumber
\end{eqnarray}
where $p=[p_x,p_y,p_z]^T \in \mathbb{R}^3$ denotes the position, $v=[v_x,v_y,v_z]^T \in \mathbb{R}^3$ represents the velocity of UAV in world frame. The UAV roll and pitch angles are defined by ${\phi,\theta}\in [-\pi,\pi]$ respectively. $R(\phi,\theta) \in SO(3)$ denotes the rotation matrix that transforms the acceleration due to thrust acting on the UAV body frame to its equivalent components into the world frame, $\mathbf{T} = [0, 0, T]^T$, $\mathbf{T} = diag(A_x, A_y, A_z)$, where $A_x$, $A_y$, $A_z \in \mathbb{R}$ represent linear damping terms, and $\mathbf{G}= [0, 0, g]^T$, with $g \in \mathbb{R}$ the Earth gravitational acceleration.
The attitude dynamics is approximated by a first-order system with time constants $\tau_\phi$ and $\tau_\theta$. The overall motion control system includes a low-level attitude controller with constant gains $K_\phi$ and $K_\theta$ to track the reference orientation commands $\phi_{ref}$ and $\theta_{ref}$ in $\mathbb{R}$. These orientation references, along with the thrust magnitude $T > 0 \in \mathbb{R}$, form the control input. This input is designed using an NMPC-based high-level position controller to achieve accurate reactive path following. The UAV dynamics described in Equation~\ref{eq:quadcopter_dynamics} undergo discretization via the forward Euler method, utilizing a sampling time $\delta_t \in \mathbb{R}^+$. 
The discretized model serves as the prediction model for the NMPC framework for each time instant $(k+j|k)$, which denotes the prediction of the time step $k+j$ produced at the time step $k$, with a prediction horizon denoted as $N$. The control design's objective is to determine an optimal sequence of $\mathbf{u}_k$ at each time step to minimize the cost function given by:
\begin{equation}\label{eq:MPC_cost_func}
        \begin{gathered}
        J=\sum_{j=1}^{N} \underbrace{ (x_{ref}-x_{k+j \mid k})^T Q_x (x_{ref}-x_{k+j \mid k})}_\text{state\space cost} \\
        +\underbrace{(u_{ref}-u_{k+j \mid k})^T Q_u (u_{ref}-u_{k+j \mid k})}_\text{input\space cost}\\
        +\underbrace{(u_{k+j \mid k}-u_{k+j-1 \mid k})^T Q_{\Delta u} (u_{k+j \mid k}-u_{k+j-1 \mid k})}_\text{input\space smoothness \space cost}
    \end{gathered}
\end{equation}
where $Q_x\in\mathbb{R}^{8\times8},Q_u ,Q_{\Delta_u} \in \mathbb{R}^{3 \times 3} $.
The $state \: cost$ penalizes the deviation of the predicted future positions with the 
reference, the $input \: cost$ is to minimize deviations from the reference control input, 
aiming to minimize the energy of the control actuation. The $input \: smoothness \:  cost$  
penalizes deviations between successive inputs, promoting smooth control action by minimizing the control rate. This optimization process aims to minimize the cost function while adhering to the system dynamics and input constraints. The input constraints are defined as $u_{min} 
\leq u \leq u_{max}$, where $u_{min}$, $u_{max} \in \mathbb{R}^3$. Additionally, constraints are imposed on the maximum rate of change of $\phi_{ref}$ and $\theta_{ref}$, bounded as $\Delta \phi_{max}$ and $\Delta \theta_{max}$, 
The optimization problem is numerically solved using PANOC with a single-shooting approach~\cite{sathya2018embedded} through the Optimization Engine (OpEn) platform \cite{opengen}.

\section{Results and Discussions} \label{Sec:Results}
The efficacy of the proposed framework is investigated by executing real-life emulation experiments in a lab environment using an in-house designed and built quad-copter. The Vicon motion-capturing system is responsible for tracking the UAV and providing odometry feedback. 
Neither the extruder arm nor the extruder tip are addressed within the scope of this article since material science and extruder technology are deliberately excluded from consideration for this specific application. However, a suitable candidate for the specific task could be considered a foam spraying mechanism similar to \cite{kovac,zhang2022aerial}
Assuming the extruder's length $l_{ex}=0.5 m$ and considering it rigidly mounted on the bottom of the UAV, its position is calculated using the data received from Vicon. Upon the issuance of a command instructing the initiation of material extrusion from the extruder, the deposition of the material is emulated by placing small spherical markers of radius $r_m = 0.3 cm$ in an offset of $0.1 m$ to account for the inherent spray effect generated by the extruder. 

\begin{figure}
    \centering
    \subfloat[Chunks of the Decomposed Mesh]{
        {\includegraphics[width=0.6\columnwidth]{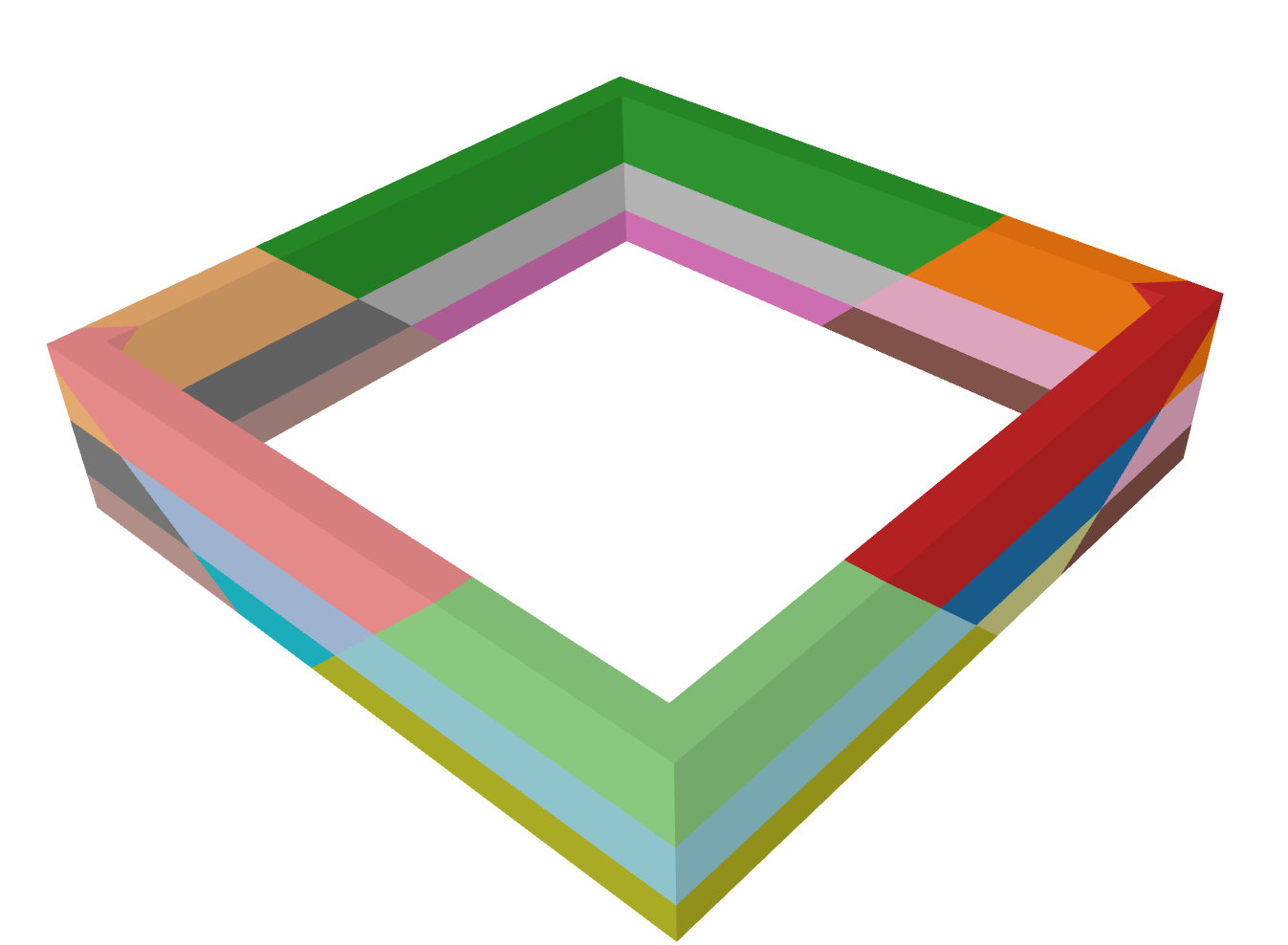} }
        \label{fig:dome_chunks}
    }
    \qquad
    \subfloat[BSP Tree $\mathcal{T}$ and priority order $\mathcal{P}_{ord}$]{{
        \includegraphics[width=\columnwidth]{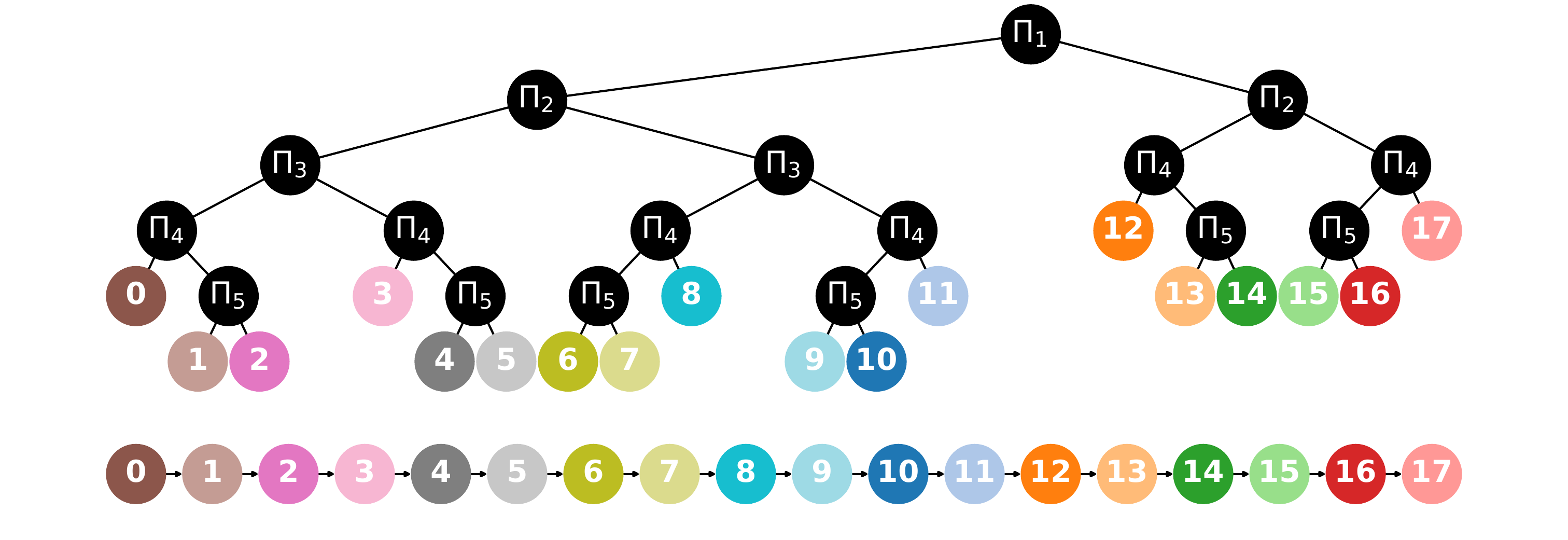} }
        \label{fig:dome_tree}
    }
    \caption{Color-coded chunking Results for the rectangle, stacked together forming the decomposed mesh(a). BSP tree $\mathcal{T}$ visualizing the cutting planes $\Pi_1,\dots, \Pi_5$ in the intermediate nodes and the chunks $C_1, \dots, C_{17}$ on the leaf nodes. The printing sequence graph $\mathcal{P}_{ord}$ is extracted at the bottom (b).}
    \label{fig:dome}
    \vspace{-4.5mm}
    
\end{figure}

The mesh used for the experiment is a $2 m$ wide, $2 m$ long and $0.5 m$ height hollow rectangle having walls with $0.1$ m width as depicted in Fig. \ref{fig:dome}.  The bounding angles of the search for the planar cuts sampling space $\mathcal{S}$ are set to $\phi_{sample}^{max} = 45 \degree$ and $\theta \in [0, 2\pi)$ while the weights of the heuristic function are designed as follows: $G_{disp}=200 $, $G_{part}=10 $ and $G_{faces} = 20$. 
The planar cuts search algorithm resulted in $5$ planes $[ \Pi_1, \Pi_5]$ as the optimal set that generated $18$ chunks. The generated chunks are shown in Fig. \ref{fig:dome_chunks} composing the whole given mesh, in a color-coded format.
The overall result of the search is illustrated in the BSP tree in Fig. \ref{fig:dome_tree} having all the cuts as intermediate nodes and all the generated color-codes chunks as leaf nodes. The final priority order calculated by the in-order traversal of the tree is shown at the bottom of Fig.\ref{fig:dome}, starting from chunk $C_0$ to $C_{17}$.

\begin{figure}
    \includegraphics[width=1.0\columnwidth]{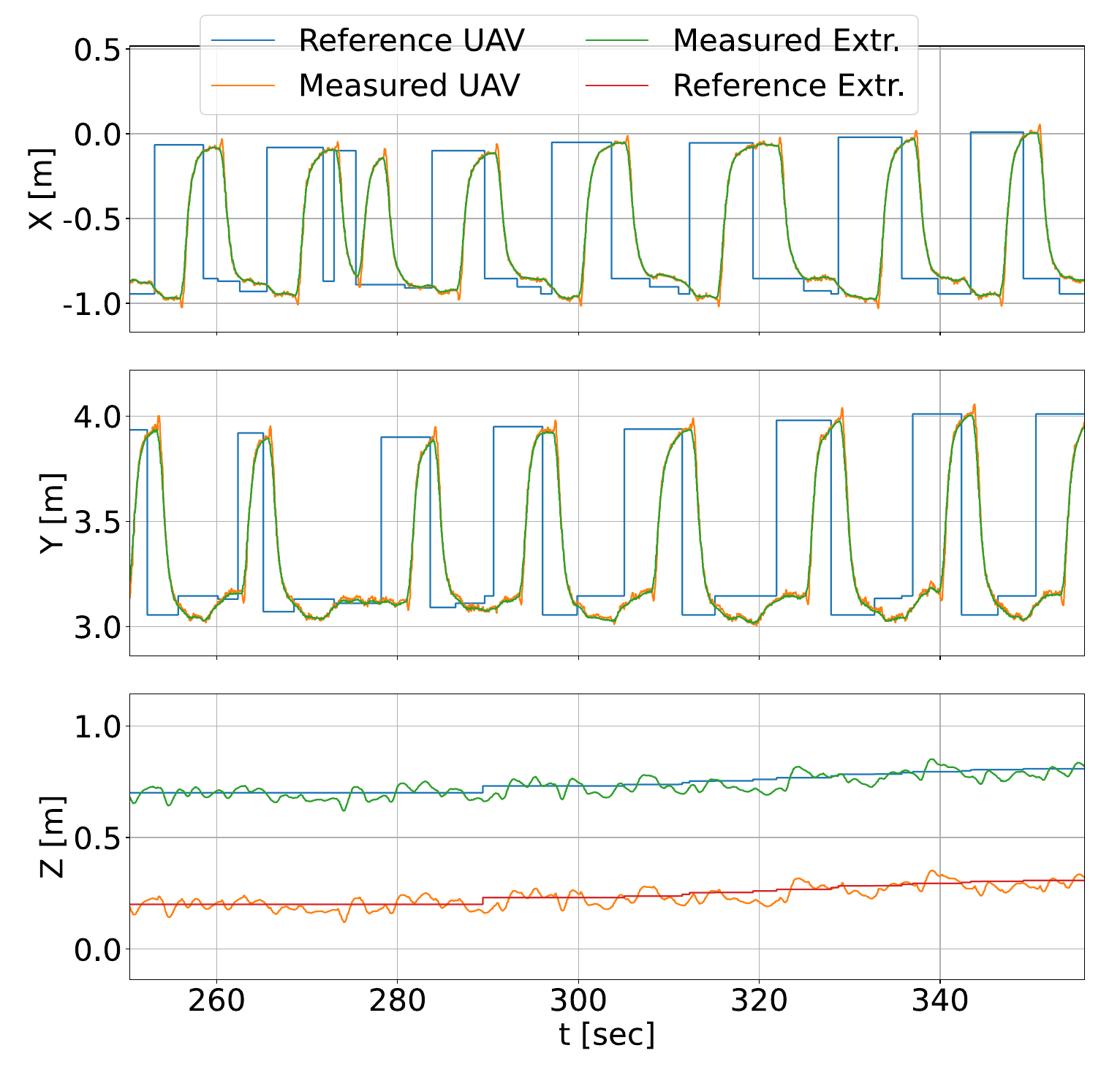}
    \caption{Reference, Measured UAV and Measured Extruder Coordinates per Axis $X$ and $Y$ from $t=271$ sec until $t=360$ sec during the emulation of the printing in the real-life UAV.}
    \label{fig:errorPerAxis}
    
\end{figure}

The model parameters as mentioned in the model \ref{eq:quadcopter_dynamics} are selected as following: $A_x = 0.1$, $A_y = 0.1$, $A_z = 0.2$ and $\tau_\phi = 0.4$, $\tau_\theta = 0.47$  and $K_\phi = K_\theta = 1$. The sampling time used is $50 \: msec$ while the prediction horizon time-steps number is set to $N=40$ corresponding to a $2$ sec prediction. The cost function weights  are set as follows $Q_x = diag(15, 15, 25, 4, 4, 4, 8, 8)$, $Q_u = diag(3, 15, 15)$ and $Q_{\Delta u} = diag(3, 15, 15)$. Additionally, the control input bounds are set to $u_{min} = [3, -0.2, -0.2 ]$ and $u_{max}= [ 15.5, 0.2, 0.2 ]$ (SI). Finally, the input rate thresholds were set to $\Delta \phi_{max} = \Delta \theta_{max} = 0.04$ rad.
\begin{figure*}    
\includegraphics[width=\textwidth]{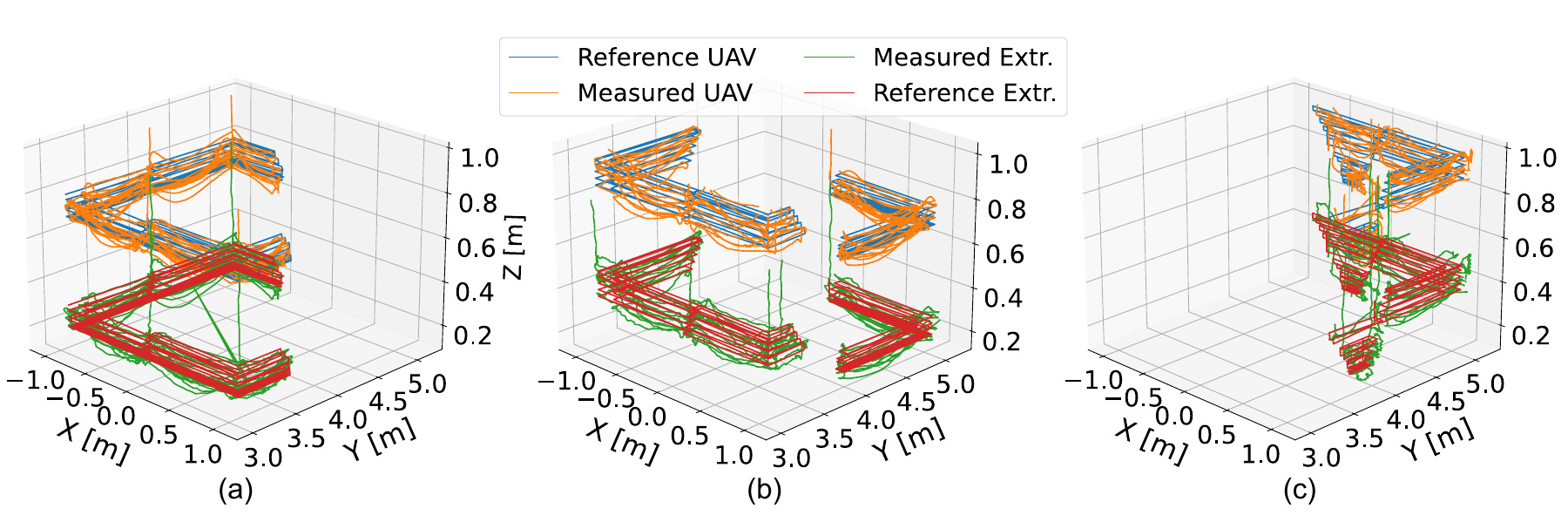}
    \centering
    \caption{Sequential 3D Plots for chunks $ C_0 - C_2$ (a), $ C_3 - C_4$ (b), $ C_7 - C_9$ (c) of Reference UAV position (blue), Measured UAV position (green), Extruder reference position (red) and calculated Extruder position (orange) during the printing procedure.}
    \label{fig:3DPlotrect}
    \vspace{-2.0mm}
\end{figure*}
\begin{figure*}[b!]
        \includegraphics[width=\textwidth]{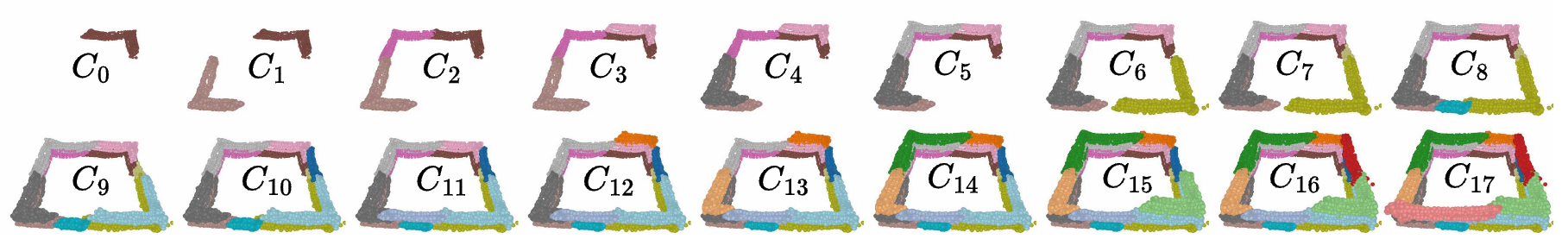}
        \caption{Sequential visualizations of the chunks $C_0$ to $C_{17}$ printing process via real-time data gathered from the motion capture system. }
        \label{fig:sequence_print}
\end{figure*}
The reference and measured position of the UAV through Vicon along with the reference position of the extruder ($l_{ex}$ m below the UAV) and its projected position as calculated from the pose of the UAV, are depicted in Fig. \ref{fig:3DPlotrect}. Observations from this plot reveal that the UAV adheres to the prescribed flight paths, executing linear movements in accordance with the slicing process.
Notably, the extruder's positional data exhibits occasional oscillations and overshoots, particularly noticeable during the initiation and conclusion of movement sequences.
This phenomenon can be attributed to the extruder's length, as slightly small changes to the roll and the pitch angle of the UAV are translated to big errors proportional to the length. Despite the aforementioned overshoots, overall the measured positions are bounded within an offset of $5$ to $6$ cm per axis. 
The data outlined are also graphically presented per axis in Figure \ref{fig:errorPerAxis}. 
The tracking of reference points within each axis exhibits a satisfactory level of performance, notwithstanding occasional extruder oscillations during movement and hovering. It is imperative to acknowledge that such oscillations are primarily stemming from model inaccuracies or potential imbalances within the aerial platform. 
Additionally, it can be observed that the position error exhibits an increased magnitude during the start and the end of each linear segment's movement. This occurrence is a predictable outcome attributable to the momentary rolling and pitching angle of the UAV during the movement initiation aiming to accurately track the reference waypoint. 
Whenever this occurs, a fast increase in the velocity is noticed. 
Not maintaining a constant speed during the execution of the printing, may lead to discontinuities and uneven material deposition throughout the traversal of the manufacturing path. Despite these variations and the presence of a deviation of the order of $6$ cm per axis from the designated waypoints, the diameter of the deposited material
is sufficiently big to compensate for it and not compromise the integrity of the whole structure as shown in Fig. \ref{fig:sequence_print}. A video with the full mission can be found in \href{https://youtu.be/gfZuYCA8jAw}{https://youtu.be/gfZuYCA8jAw}
%
\section{Conclusions} \label{sec:conclusions}
This article introduces a novel aerial 3D printing framework and demonstrates it through experimental emulation.
In the selected evaluation scenario a rectangular $2$x$2$x$0.5$ m wall has been emulated for printing carried out by deploying real-life tracked UAVs executing the planned print task in the adequate order imposed by the decomposition process.
This framework has the potential to revolutionize additive manufacturing by enabling large-scale structures through its capability to divide complex 3D models into manageable components, enabling distributed 3D printing. 
However, it's in the early stages, and further validation is needed. The use of experimental emulation helps identify gaps between simulation and reality, providing a foundation for addressing challenges in 3D printing.

\clearpage
\bibliographystyle{IEEEtran}
\bibliography{sample}
\end{document}